# Network-Level Vehicle Delay Estimation at Heterogeneous Signalized Intersections

**Xiaobo Ma, Ph.D. (Corresponding author)**
Pima Association of Governments
1 E Broadway Blvd., Ste. 401
Tucson, AZ 85701
Email: XMa@pagregion.com

**Hyunsoo Noh, Ph.D.**
Pima Association of Governments
1 E Broadway Blvd., Ste. 401
Tucson, AZ 85701
Email: HNoh@pagregion.com

**James Tokishi**
Pima Association of Governments
1 E Broadway Blvd., Ste. 401
Tucson, AZ 85701
Email: JTokishi@pagregion.com

**Ryan Hatch**
Pima Association of Governments
1 E Broadway Blvd., Ste. 401
Tucson, AZ 85701
Email: RHatch@pagregion.com

# ABSTRACT

Accurate vehicle delay estimation is essential for evaluating the performance of signalized intersections and informing traffic management strategies. Delay reflects congestion levels and affects travel time reliability, fuel use, and emissions. Machine learning (ML) offers a scalable, cost-effective alternative; However, conventional models typically assume that training and testing data follow the same distribution, an assumption that is rarely satisfied in real-world applications. Variations in road geometry, signal timing, and driver behavior across intersections often lead to poor generalization and reduced model accuracy. To address this issue, this study introduces a domain adaptation (DA) framework for estimating vehicle delays across diverse intersections. The framework separates data into source and target domains, extracts key traffic features, and fine-tunes the model using a small, labeled subset from the target domain. A novel DA model, Gradient Boosting with Balanced Weighting (GBBW), reweights source data based on similarity to the target domain, improving adaptability. The framework is tested using data from 57 heterogeneous intersections in Pima County, Arizona. Performance is evaluated against eight state-of-the-art ML regression models and seven instance-based DA methods. Results demonstrate that the GBBW framework provides more accurate and robust delay estimates. This approach supports more reliable traffic signal optimization, congestion management, and performance-based planning. By enhancing model transferability, the framework facilitates broader deployment of machine learning techniques in real-world transportation systems.

# 1. INTRODUCTION

Vehicle delay estimation is essential for evaluating the operational efficiency of signalized intersections and overall roadway performance(Ma, Karimpour, et al., 2023c; Yang et al., 2024). Delay is a direct indicator of traffic congestion and driver experience, influencing travel time reliability, fuel consumption, and emissions. Accurate delay estimation enables transportation agencies to identify underperforming intersections, prioritize signal timing adjustments, and implement data-driven strategies to improve traffic flow(Ma, Cottam, et al., 2023; Wu et al., 2019). It also supports performance-based planning, congestion management, and infrastructure investment decisions (Ma, Karimpour, et al., 2023a). In the context of intelligent transportation systems, real-time or predictive delay estimation contributes to adaptive signal control and proactive traffic management, ultimately enhancing mobility and safety for all road users (Ma, Noh, et al., 2024; Z. Wang et al., 2025; Y. Xu et al., 2025).

Recent advancements in machine learning and data-driven approaches have offered promising approaches for estimating vehicle delays. Several studies have explored the application of artificial intelligence and machine learning techniques for estimating vehicle delay at signalized intersections. One study employed Fuzzy Logic (FL) and Artificial Neural Networks (ANN), to model vehicle delay under various traffic conditions. Two models were developed: the Neuro-Fuzzy Delay Estimation (NFDE) model and the Artificial Neural Network Delay Estimation (ANNDE) model. The data used for model development and validation were obtained from ten signalized intersections(Sazi Murat, 2006). Gene Expression Programming (GEP), an artificial intelligence technique, has also been used to construct models capable of estimating vehicle delay. The models were validated using data collected from 18 signalized intersections(Bagdatli, 2020). Another study proposed estimation models based on four distinct machine learning algorithms—Support Vector Regression (SVR), Random Forest (RF), k-Nearest Neighbors (kNN), and Extreme Gradient Boosting (XGBoost). The models were evaluated using data collected from 12 signalized intersections(Bagdatli & Dokuz, 2021). Additionally, one study aimed to efficiently and accurately estimate vehicle delay using the Artificial Bee Colony (ABC) algorithm and the Flower Pollination Algorithm (FPA), demonstrating the potential of nature-inspired heuristics in traffic delay evaluation(Korkmaz & Akgüngör, 2020). In a more recent study, the authors provide an in-depth examination of various machine learning models, including Support Vector Regression (SVR), k-Nearest Neighbors (kNN), Artificial Neural Networks (ANN), Random Forests (RF), and Decision Trees (DT), demonstrating the effectiveness of these techniques for estimating vehicle delay(Ranpura et al., 2024).

Machine learning methods offer a scalable and cost-effective alternative to traditional approaches, enabling deployment across multiple intersections with minimal additional infrastructure (Cottam et al., 2024; Ma, Karimpour, et al., 2023b; Ma, Noh, et al., 2025). However, despite their advantages, conventional machine learning models typically assume that the training and testing datasets share identical data distributions—an assumption that rarely holds in real-world traffic environments. Traffic patterns can vary widely between intersections due to differences in road geometry, signal timing configurations, and local driver behavior (Ma, 2022; Ma, Karimpour, et al., 2024). Such domain discrepancies limit the generalizability and accuracy of traditional machine learning models when applied to previously unseen intersections (Luo et al., 2022; Ma et al., 2020; Z. Zhang et al., 2024). Similar concerns have been raised in other domains where artificial intelligence and generative AI technologies are increasingly integrated into knowledge production and decision-making. For instance, research in applied linguistics and composition studies has shown that while AI tools expand opportunities for multimodal expression, authorship, and efficiency, their effectiveness depends heavily on context-specific constraints, genre expectations, and users' perceptions of credibility and trust (Diaz et al., n.d.; Hakim et al., 2024; Tan et al., 2024; Tan, Wang, et al., 2025; Tan, Xu, et al., 2025b, 2025a; Tan & Xu, 2024; C. Wang et al., 2025; W. Xu, 2017, 2023a, 2023c, 2023b, 2024, 2025; W. Xu & Jia, 2025; W. Xu & Tan, 2024, 2025). These insights highlight that, across disciplines, AI models must be critically adapted to local conditions rather than applied under the assumption of uniform data or interpretive practices.

To address domain discrepancies and enhance model generalizability, the concept of domain adaptation (DA) has gained considerable attention in recent years. DA is an advanced machine learning approach that transfers knowledge from related tasks or source domains to improve performance on a target task. By relaxing the restrictive assumption that source and target data distributions must be identical, DA enables pre-trained models to adapt to new domains with limited target data, thereby improving prediction accuracy and robustness(Ma, Karimpour, et al., 2025). In traffic management, DA leverages data-rich cities or intersections (source domains) to infer traffic patterns in data-scarce locations (target domains). For example, Yao et al. (2023) developed a traffic prediction model tailored for data-sparse road networks by transferring knowledge from data-abundant networks. Their approach integrates spatial-temporal graph convolutional networks with adversarial domain adaptation to extract domain-invariant features for effective knowledge transfer(Yao et al., 2023). Moreover, Mo and Gong (2023) proposed a Cross-city Multi-Granular Adaptive Transfer Learning method that utilizes limited target city data. This model applies meta-learning to initialize training across multiple source cities and extracts multi-granular regional features. An Adaptive Transfer module, incorporating Spatial-Attention and Multi-head Attention mechanisms, selectively transfers the most relevant features to the target domain (Mo & Gong, 2022). More recently, Li et al. (2024) introduced a macroscopic fundamental diagram (MFD)-guided transfer learning framework that identifies and transfers domain-invariant traffic flow patterns to tackle challenges such as data scarcity and dataset shifts. By employing an MFD similarity measure, the method identifies transferable patterns and integrates this measure into domain adversarial pre-training for enhanced adaptability (Li et al., 2024).

Although existing studies have demonstrated the effectiveness of DA in traffic prediction and estimation tasks, to the best of the authors' knowledge, no DA model has yet been applied specifically to vehicle delay estimation. To address the aforementioned limitations and bridge the existing research gaps, this research proposes a DA framework for estimating vehicle delay. The framework begins with separating the source and target domains. Each domain includes traffic data and corresponding vehicle delay values. In the second step, relevant features are extracted from both domains to capture key traffic variables that influence vehicle delay. This process converts raw data into structured variables, enabling more effective model training. A small portion of labeled data from the target domain is then used to fine-tune the model. This step allows the model—initially trained on source domain data—to better adapt to the unique traffic patterns of the target domain. Next, a domain adaptation model, GBBW, is applied to estimate vehicle delays across all target intersections. During training, data instances more similar to the fine-tuning subset are given higher weights, allowing the model to focus on patterns most relevant to the target domain. The final output is a set of vehicle delay predictions, offering actionable insights for traffic signal optimization, congestion management, and broader transportation planning.

The proposed framework develops scene-specific models by employing a DA approach. DA addresses a common challenge in machine learning: the mismatch between the data distributions of the training (source) and testing (target) domains. By storing knowledge learned from previously trained models and reapplying it to similar but distinct scenarios, DA enables the model to generalize effectively across different contexts. In this framework, the DA model relaxes the assumption that the source and target domains share identical distributions, allowing for accurate estimation of vehicle delays across intersections with varying traffic patterns and characteristics. The framework's performance is validated using data from 57 heterogeneous intersections in Pima County, Arizona.

This research makes several key contributions to the field of traffic management and vehicle delay estimation:

- **Novel DA framework for vehicle delay estimation:** This study introduces a novel domain adaptation (DA) framework for vehicle delay estimation—marking the first known application of DA in this context. The proposed Gradient Boosting with Balanced Weighting (GBBW) model extends traditional Gradient Boosting by incorporating domain adaptation through balanced weighting.
- **Comprehensive comparison with state-of-the-art models:** The proposed DA framework is systematically evaluated against leading models to demonstrate its effectiveness and to identify potential limitations.
- **Scene-specific, scalable, and cost-effective modeling approach:** Leveraging traffic controller event data, the framework offers an intersection-specific, scalable, and cost-efficient solution that adapts to the dynamic nature of urban traffic.
- **Improved data efficiency through cross-domain learning:** By transferring knowledge from related intersections, the framework enhances data efficiency and enables accurate vehicle delay estimation even when target domain data are limited.

In conclusion, the proposed DA framework offers a significant advancement in vehicle delay estimation. By addressing the shortcomings of traditional models and leveraging the capabilities of domain adaptation, this study presents a robust, efficient, and scalable solution to support modern urban traffic management.

## 2. METHODOLOGY

### 2.1. Research Framework

The purpose of this study is to estimate vehicle delay for intersections. This goal can be achieved through training models on a known intersection (henceforth referred to as "source domain") and transferring the well-trained models to estimate vehicle delay for a new intersection (henceforth referred to as "target domain"). The proposed framework for estimating vehicle delay using DA consists of several key components, as depicted in Figure 1. These components are designed to systematically process and utilize data from both source and target domains to estimate vehicle delay accurately.

In the first step, isolating domains is performed. Let's assume the entire data set of the source and target domain intersections are denoted as $\mathbb{D}_S$ and $\mathbb{D}_T$, respectively; in this case, $\mathbb{D}_S = (\mathbb{X}_S, \mathbb{Y}_S)$ and $\mathbb{D}_T = (\mathbb{X}_T, \mathbb{Y}_T)$. Both $\mathbb{D}_S$ and $\mathbb{D}_T$ have two parts: data instances $\mathbb{X}_S$ and $\mathbb{X}_T$ as well as labels $\mathbb{Y}_S$ and $\mathbb{Y}_T$. Assume $\mathbb{X}_S$ and $\mathbb{X}_T$ can be represented as $n$ by $p$ matrices, where $n$ denotes the total number of observations (or records), and $p$ denotes the total number of variables in the dataset. The data instances $\mathbb{X}_S$ and $\mathbb{X}_T$ serve as tentative inputs for model training, while $\mathbb{Y}_S$ and $\mathbb{Y}_T$ represent the model outputs, which correspond to vehicle delays in this study.

In the second step, both source and target domain data undergo feature extraction processes to identify relevant traffic variables (e.g., signal and detection events) that contribute to vehicle delay patterns. Feature extraction is a critical step as it transforms raw traffic data into structured information that can be utilized in subsequent steps. Let $D_S$ represent data extracted from the source domain intersections and $D_T$ represent data extracted from the target domain intersection. Similarly, $D_S = (\mathcal{X}_S, \mathcal{Y}_S)$ and $D_T = (\mathcal{X}_T, \mathcal{Y}_T)$. $\mathcal{X}_S$ and $\mathcal{X}_T$ contain all the variables extracted from the second step. Assume the number of variables extracted is $q$, then both $\mathcal{X}_S$ and $\mathcal{X}_T$ become $n$ by $q$ matrices.

Let $\mathcal{X}_{T-S}$ and $\mathcal{Y}_{T-S}$ represent small subsets of data extracted from the target domain. The proposed framework utilizes these subsets to fine-tune the model during training. This fine-tuning step enables the model, initially trained on source domain data, to adapt and infer traffic patterns more accurately for the target domain.

Next, a DA model named GBBW is deployed to estimate vehicle delay for all the intersections in the target domain. In the model training process, labeled data instances that are similar to the $\mathcal{X}_{T-S}$ and $\mathcal{Y}_{T-S}$ are assigned higher weights, while less similar instances receive lower weights. By emphasizing these more relevant instances, the model focuses on information that better aligns with the target domain, thereby improving the accuracy and effectiveness of the trained regressor. The final output of the framework is a set of vehicle delay estimates. These predictions provide valuable insights for traffic management authorities, supporting signal optimization, congestion mitigation, and strategic planning.

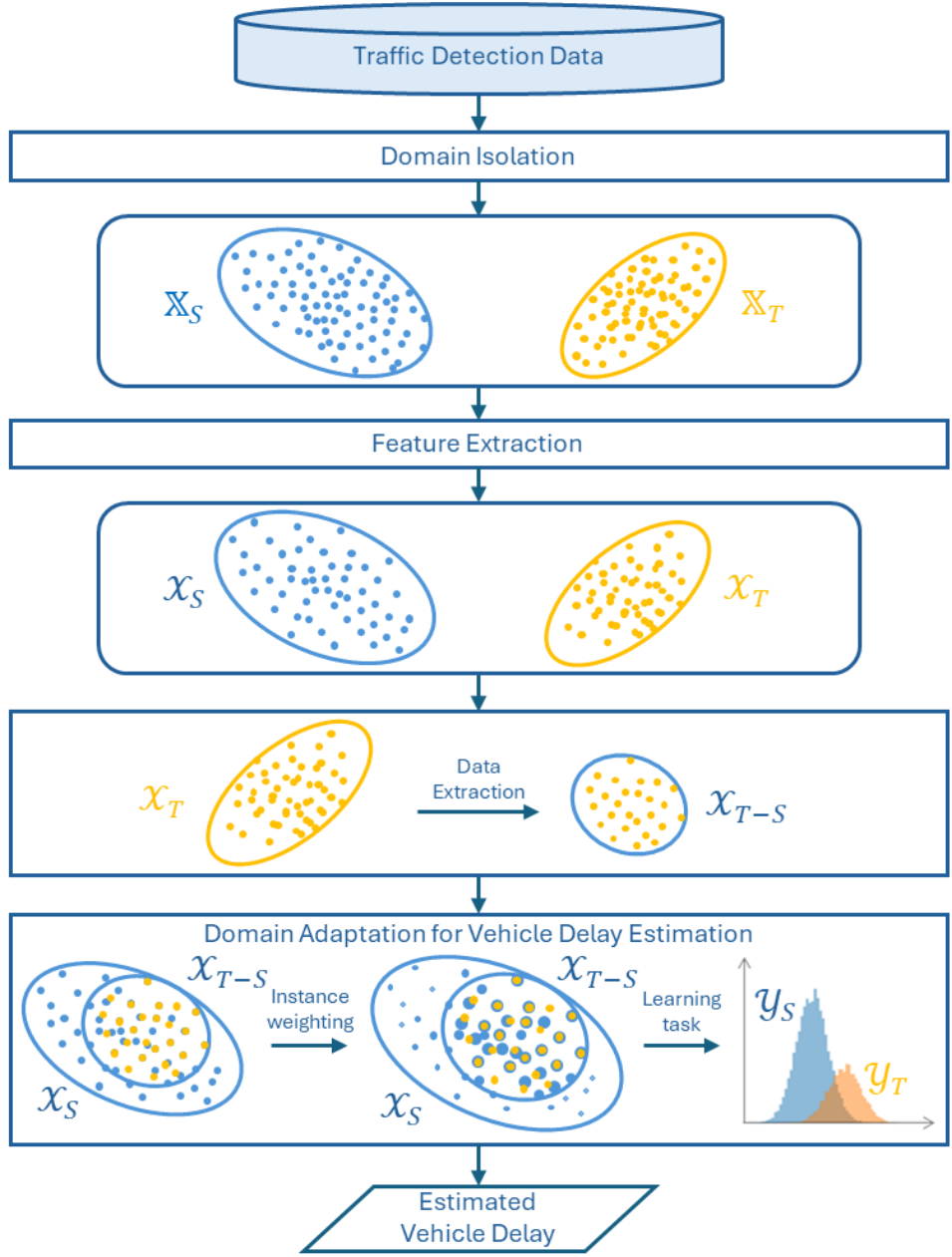

**Figure 1 Research framework**

## 2.2. Gradient Boosting

Gradient boosting is used in supervised learning to find a function $\hat{F}(x)$ that best predicts an output variable $y$ from input variables $x$. This is achieved by introducing a loss function $L(y, F(x))$, and minimizing its expected value:

$$\hat{F} = \arg\min_{F} \mathbb{E}_{x,y}[L(y, F(x))] \tag{1}$$

The method approximates $\hat{F}(x)$ as a sum of $M$ simpler functions $h_m(x)$ from some class $\mathcal{H}$, called base or weak learners:

$$\hat{F}(x) = \sum_{m=1}^{M} \gamma_m h_m(x) + \text{const.} \qquad (2)$$

Given a training set of $\{(x_1, y_1), \ldots, (x_n, y_n)\}$, the algorithm aims to minimize the empirical risk (average loss on the training set). It starts with a constant function $F_0(x)$ and builds the model incrementally:

$$F_0(x) = \arg\min_{\gamma} \sum_{i=1}^{n} L(y_i, \gamma) \qquad (3)$$

$$F_m(x) = F_{m-1}(x) + \left(\arg\min_{h_m \in \mathcal{H}} \left[\sum_{i=1}^{n} L(y_i, F_{m-1}(x_i) + h_m(x_i))\right]\right)(x) \qquad (4)$$

Since finding the optimal $h_m$ at each step is computationally infeasible, a simplified approach is used. The algorithm applies steepest descent, moving a small amount $\gamma$ in the negative gradient direction of the loss function:

$$F_m(x) = F_{m-1}(x) - \gamma \sum_{i=1}^{n} \nabla_{F_{m-1}} L(y_i, F_{m-1}(x_i)) \qquad (5)$$

This process continues for $M$ iterations, gradually improving the model's predictive power by combining multiple weak learners into a strong predictor (Friedman, 2001).

**2.3. Gradient Boosting with Balanced Weighting**

Balanced Weighting is an instance-based domain adaptation technique that assigns weights to samples to account for data distribution differences between source and target domains. It balances the source distribution by maximizing the weights of samples similar to the target distribution and minimizing irrelevant ones. This improves model performance on target tasks, ensures better generalization, and effectively handles domain shifts (De Mathelin et al., 2022). This study proposes GBBW as an extension of Gradient Boosting by integrating the Balanced Weighting technique for domain adaptation. This approach builds on the strengths of Gradient Boosting, a powerful ensemble learning method, and incorporates Balanced Weighting to address distribution differences between source and target domains. By emphasizing samples from the source domain that align closely with the target distribution, it effectively handles domain shifts while leveraging gradient boosting's strong predictive performance. This method enhances model generalization to target tasks and improves accuracy in scenarios with distribution differences between source and target domains. In the GBBW, the ratio parameter, $\alpha$, controls the balance between the influence of source and target data in the loss function. This approach enables the model to effectively leverage information from both domains, enhancing its generalization capability across diverse datasets. By dynamically adjusting the contributions of source and target data, the method facilitates domain adaptation by weighting the importance of each domain based on the value of $\alpha$.

During the initialization step, both source and target data are incorporated, with weights of $1 - \alpha$ and $\alpha$, respectively. Pseudo-residuals are calculated separately for the source and target data, and the base learner is trained on a combined set of pseudo-residuals from both domains, weighted by $1 - \alpha$ for source samples and $\alpha$ for target samples. The multiplier computation further accounts for losses from both source and target data, applying weights of $1 - \alpha$ and $\alpha$ accordingly. This approach allows the Gradient Boosting algorithm to learn from both source and target domains, effectively balancing their contributions through the $\alpha$ parameter. By assigning appropriate importance to each domain, it significantly aids in domain adaptation tasks during the learning process (De Mathelin et al., 2022). To simplify the notation, the source domain data is denoted as $\{(x_i, y_i)\}_{i=1}^{n_1}$ (source), which is $D_S$ in this study. The target domain data is denoted as $\{(x_j, y_j)\}_{j=1}^{n_2}$ (target), which corresponds to $\mathcal{X}_{T-S}$ and $\mathcal{Y}_{T-S}$. $L(\cdot)$ is a loss function.

---

**Input** Two training sets $\{(x_i, y_i)\}_{i=1}^{n_1}$ (source) and $\{(x_j, y_j)\}_{j=1}^{n_2}$ (target), a differentiable loss function $L(y, F(x))$, number of iterations $M$, and a weighting parameter $\alpha$.

**Initialize model with a constant value:**

$$F_0(x) = \arg\min_{\gamma}\left((1-\alpha)\sum_{i=1}^{n_1} L(y_i, \gamma) + \alpha \sum_{j=1}^{n_2} L(y_j, \gamma)\right)$$

**For** $m = 1$ to $M$:
1. Compute pseudo-residuals for both training sets:

$$r_{im} = -\left[\frac{\partial L(y_i, F(x_i))}{\partial F(x_i)}\right]_{F(x)=F_{m-1}(x)} \quad \text{for } i = 1, \ldots, n_1$$

$$r_{jm} = -\left[\frac{\partial L(y_j, F(x_j))}{\partial F(x_j)}\right]_{F(x)=F_{m-1}(x)} \quad \text{for } j = 1, \ldots, n_2$$

2. Fit a base learner $h_m(x)$ to pseudo-residuals:
   Train it using the combined training set $\{(x_i, r_{im})\}_{i=1}^{n_1} \cup \{(x_j, r_{jm})\}_{j=1}^{n_2}$

3. Compute multiplier $\gamma_m$ by solving the following optimization problem:

$$\gamma_m = \arg\min_{\gamma}\left((1-\alpha)\sum_{i=1}^{n_1} L(y_i, F_{m-1}(x_i) + \gamma h_m(x_i)) + \alpha \sum_{j=1}^{n_2} L(y_j, F_{m-1}(x_j) + \gamma h_m(x_j))\right)$$

4. Update the model:

$$F_m(x) = F_{m-1}(x) + \gamma_m h_m(x)$$

**Output** $F_M(x)$

---

## 3. EXPERIMENTS

### 3.1. Data Description

In the Pima County region, two major traffic detection systems, Miovision and MaxView, are used for actuated signal control and event-based data collection. The Miovision sensors, configured by the Pima County Department of Transportation, collect performance measures from approximately 100 signalized intersections. These sensors provide simple delay, arrival-on-green (AoG), arrival-on-red (AoR), and split failure data through the TrafficLink portal. Miovision-based traffic performance measures have been collected since 2021 and are used to assess traffic conditions.

As shown in Figure 2, based on data availability, 57 intersections controlled by the Miovision system were selected for this study. The data was aggregated into one-hour intervals from January 7, 2021, to November 30, 2021.

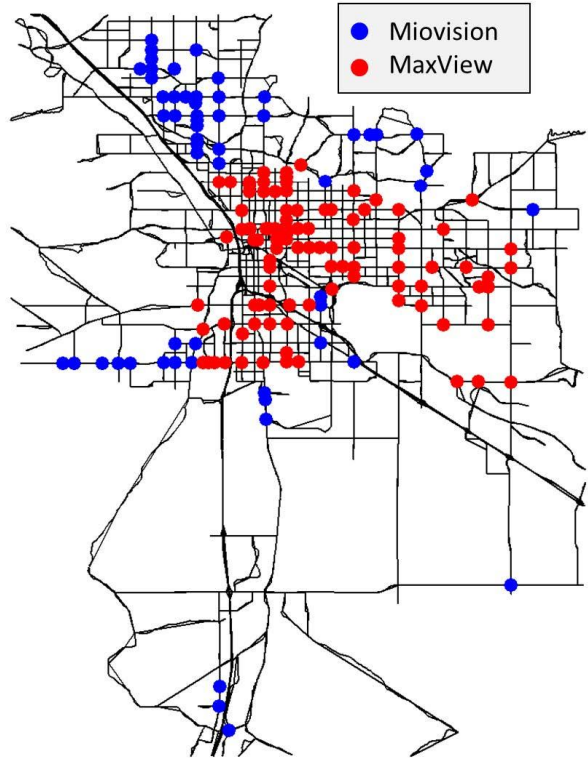

**Figure 2 Locations of the study intersections**

In a typical Miovision sensor configuration, through movements in all four directions are equipped with both presence and advance detectors to cover multiple through lanes. Presence detectors are implemented as long loops, while advance detectors are short loops. For left-turn movements in all four directions, only presence detectors are configured, with a single detector often covering multiple left-turn lanes.

It should be noted that the delay performance metric provided by the Miovision system is "simple stop delay" rather than control delay. This metric is defined as the time between stop bar detector actuation during the red phase and the onset of the green phase.

### 3.2. Input Variables

Since signal and detection events are closely related to performance measures, the first step involves processing event-based data to derive the input variables required for the proposed estimation model.

- **Occupancy time:** The time difference between a vehicle triggering and leaving a detector.

- **Waiting time:** The time difference from the arrival of the first vehicle at the intersection during the red phase until the green light is activated.

In addition, the number of detection events is an important variable indicating traffic conditions. The signal status also has a significant impact on the three detection-related variables above. Therefore, signal status should be considered when extracting these variables.

All signal statuses are categorized into three groups based on the combination of signal status when the detector is activated and deactivated:

- **Red-to-Green:** Vehicles trigger the detector during the red phase and exit during the green phase.

- **Red-to-Red:** Vehicles trigger the detector during the red phase and exit during the red phase.

- **Green-to-Green:** Vehicles trigger the detector during the green phase and exit during the green phase.

In addition to variables extracted from event-based data, other relevant input variables include speed limit, hour of the day, number of lanes, and presence of shared lanes.

### 3.3. Baseline Models

Eight state-of-the-art machine learning regression models were selected as baseline models to evaluate the feasibility of the proposed framework for estimating vehicle delays. These models include Support Vector Regression (SVR) (Drucker et al., 1996), K-Nearest Neighbors (KNN) (Cover & Hart, 1967), Multi-Layer Perceptron (MLP) (Rumelhart et al., 1986), Random Forest (RF)(Breiman, 2001), Adaptive Boosting (AdaBoost) (Freund & Schapire, 1995), Categorical Boosting (CatBoost) (Prokhorenkova et al., 2018), eXtreme Gradient Boosting (XGBoost) (Chen & Guestrin, 2016), and Light Gradient Boosting Machine (LightGBM) (Ke et al., 2017). Additionally, seven state-of-the-art instance-based domain adaptation methods were included as baseline models for comparison: Importance Weighting Classifier (IWC) (Bickel et al., 2007), Transfer AdaBoost for Regression (TrAdaBoostR2)(Pardoe & Stone, 2010), Two Stage Transfer AdaBoost for Regression (TwoStageTrAdaBoostR2)(Pardoe & Stone, 2010), Relative Unconstrained Least-Squares Importance Fitting (RULSIF) (Yamada et al., 2011), Unconstrained Least-Squares Importance Fitting (ULSIF)(Kanamori et al., 2009), Kullback–Leibler Importance Estimation Procedure (KLIEP)(Sugiyama et al., 2007), and Kernel Mean Matching (KMM)(Huang et al., 2006).

A comprehensive grid search was conducted to systematically optimize the hyperparameters for both the baseline and proposed models. Experiments were conducted using Python 3.10.9 on a system with a 12th Gen Intel Core i7-12700KF CPU.

### 3.4. Measurements of Effectiveness

Mean Absolute Percentage Error (MAPE), Mean Absolute Error (MAE), and Root Mean Square Error (RMSE) are three common criteria used to evaluate and compare prediction methods (Luo et al., 2022). MAPE and MAE are used to measure the overall errors of the estimation results and RMSE is used to quantify the stability of the estimation results (W. Zhang et al., 2022). These three criteria are employed as performance metrics for comparison in this study and are defined below:

$$\text{MAPE} = 100\% * \frac{1}{N}\sum_{k=1}^{N}\left|\frac{y(k)-\hat{y}(k)}{y(k)}\right| \tag{6}$$

$$\text{MAE} = \frac{1}{N}\sum_{k=1}^{N}|y(k)-\hat{y}(k)| \tag{7}$$

$$\text{RMSE} = \sqrt{\frac{1}{N}\sum_{k=1}^{N}(y(k)-\hat{y}(k))^2} \tag{8}$$

where $y(k)$ is the actual value at time interval $k$ and $\hat{y}(k)$ is the corresponding predicted value. $N$ is the size of the testing data set (total number of time intervals).

### 3.5. Vehicle Delay Estimation Results

During the model training process, a dataset consisting of 57 distinct intersections was employed. In each iteration, one intersection was designated as the target domain, while the remaining 56 served as the source domain. This process was repeated 57 times, ensuring that each intersection was evaluated as the target domain once, thereby enabling a comprehensive assessment of the model.

Performance metrics were computed for each iteration to thoroughly evaluate the model's effectiveness and robustness across all scenarios, ensuring a reliable measure of its generalization capability. For domain adaptation models, in each iteration, 72 samples (3 days × 24 samples/day) extracted from the target domain were used for left-turn vehicle delay estimation, while 96 samples (4 days × 24 samples/day) from the target domain were required for through movement vehicle delay estimation.

Table 1 presents a comparative analysis of various models used to estimate vehicle delay for left-turn and through movements, evaluated using MAPE, MAE, and RMSE. For the left-turn movement, models such as RF, XGBoost, LightGBM, and GBBW demonstrated superior performance with notably lower error values. Among them, GBBW achieved the best results (MAPE: 10.54%, MAE: 5.37, RMSE: 7.47), indicating high estimation accuracy. Domain adaptation methods like IWC, TrAdaBoostR2, and TwoStageTrAdaBoostR2 exhibited relatively higher MAPE values (>40%), although their MAE and RMSE were moderately low, suggesting they may capture central tendency but not distributional variance effectively. In the through movement, a similar trend was observed, with GBBW again yielding the lowest errors (MAPE: 12.63%, MAE: 2.40, RMSE: 3.30). MLP, RF, CatBoost, XGBoost also performed competitively, whereas domain adaptation techniques (e.g., IWC, TrAdaBoostR2) resulted in significantly higher MAPE values, indicating limited effectiveness for estimating vehicle delay in through movements. Overall, GBBW consistently provided the most accurate and reliable vehicle delay estimates across both movement types.

**Table 1 Comparison of MAPE, MAE, and RMSE for Estimating Vehicle Delay Across Different Movements**

| Movement | Model | MAPE (%) | MAE | RMSE |
|---|---|---|---|---|
| Left-turn | SVR | 30.84 | 12.69 | 16.75 |

|  | Model | | | |
|---|---|---|---|---|
|  | KNN | 29.26 | 12.88 | 16.96 |
|  | MLP | 21.58 | 7.97 | 12.85 |
|  | RF | 11.26 | 5.59 | 7.72 |
|  | AdaBoost | 23.60 | 9.23 | 11.48 |
|  | CatBoost | 13.15 | 5.90 | 8.18 |
|  | XGBoost | 11.62 | 5.65 | 8.02 |
|  | LightGBM | 11.30 | 5.47 | 7.58 |
|  | IWC | 48.02 | 8.54 | 12.25 |
|  | TrAdaBoostR2 | 43.78 | 7.84 | 11.10 |
|  | TwoStageTrAdaBoostR2 | 42.10 | 7.16 | 10.55 |
|  | RULSIF | 17.04 | 7.47 | 10.49 |
|  | ULSIF | 17.03 | 7.47 | 10.49 |
|  | KLIEP | 16.63 | 7.37 | 10.28 |
|  | KMM | 16.12 | 7.49 | 10.64 |
|  | GBBW | ***10.54*** | ***5.37*** | ***7.47*** |
| Through | SVR | 31.34 | 5.21 | 6.96 |
|  | KNN | 33.33 | 5.71 | 7.55 |
|  | MLP | 16.07 | 2.75 | 3.79 |
|  | RF | 14.75 | 2.70 | 3.55 |
|  | AdaBoost | 35.83 | 4.86 | 6.00 |
|  | CatBoost | 15.62 | 2.73 | 3.64 |
|  | XGBoost | 14.47 | 2.63 | 3.47 |
|  | LightGBM | 15.15 | 2.68 | 3.53 |
|  | IWC | 62.72 | 3.66 | 5.95 |
|  | TrAdaBoostR2 | 64.42 | 3.43 | 4.77 |
|  | TwoStageTrAdaBoostR2 | 63.31 | 4.60 | 10.21 |
|  | RULSIF | 39.25 | 4.00 | 5.83 |
|  | ULSIF | 39.24 | 4.00 | 5.83 |
|  | KLIEP | 35.15 | 3.85 | 5.63 |
|  | KMM | 20.40 | 3.11 | 4.41 |
|  | GBBW | ***12.63*** | ***2.40*** | ***3.30*** |

Figure 3 presents a box plot comparing the MAPE across various machine learning and domain adaptation models for two movement types: left-turn and through movements. Overall, the proposed GBBW consistently outperforms traditional machine learning models (e.g., SVR, KNN, MLP, RF), boosting-based models (e.g., XGBoost, CatBoost, LightGBM), and domain adaptation models (e.g., RULSIF, ULSIF, KLIEP, KMM) in both movement types, with notably lower MAPE values and narrower interquartile ranges. Among traditional models, MLP and RF perform relatively well for left-turn movements but show greater error variability in through movements. Domain adaptation models such as IWC, TrAdaBoostR2, and TwoStageTrAdaBoostR2 display higher MAPE and variability, particularly in the through movement category, indicating potential limitations in generalizability across heterogeneous intersections. The through movements exhibit higher overall MAPE compared to left-turns across most models, suggesting greater complexity or inconsistency in delay estimation for this movement type. These results highlight the effectiveness of GBBW in achieving robust and accurate delay estimation at the network level, especially in the presence of heterogeneous intersection characteristics.

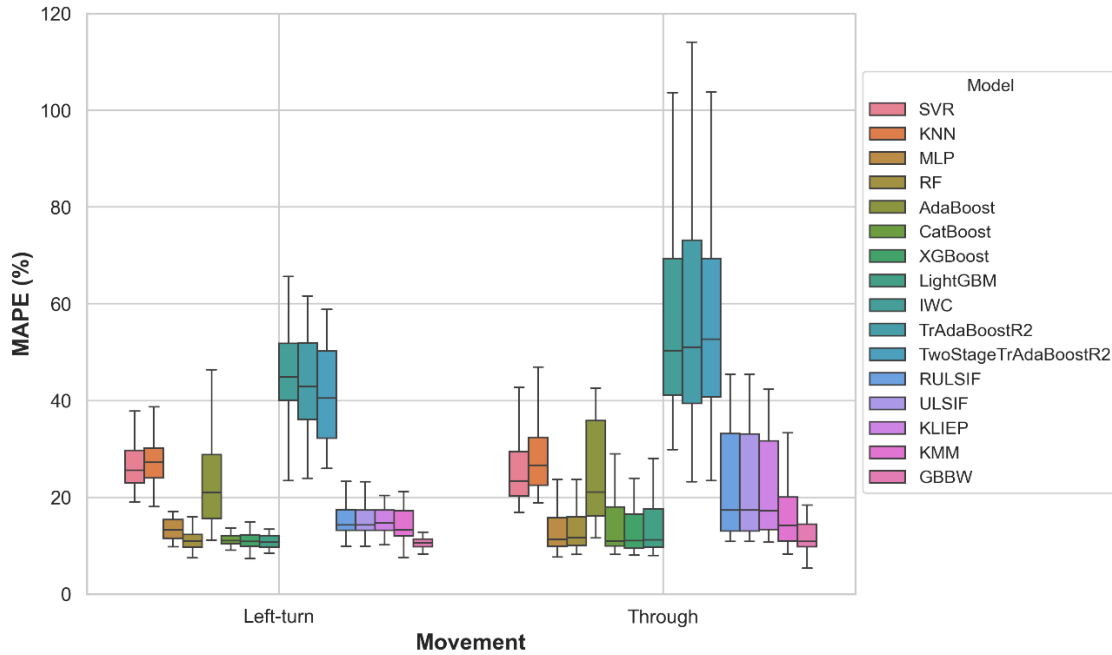

**Figure 3 MAPE comparison across different models and movement types**

Figure 4 displays the comparison of MAE across various modeling approaches for left-turn and through movements at signalized intersections. Overall, the proposed GBBW model yields the lowest MAE values and reduced variability compared to traditional machine learning models like SVR, KNN, and MLP, particularly for the through movement. Traditional models such as SVR and KNN exhibit the highest error levels, indicating limited effectiveness in generalizing across heterogeneous intersections. Boosting models including XGBoost, CatBoost, and LightGBM perform relatively well, with moderate error levels and narrower interquartile ranges, especially for the left-turn movement. The domain adaptation models (IWC, TrAdaBoostR2, and TwoStageTrAdaBoostR2) show moderate performance but with higher variability. Across both movement types, the through movement tends to result in slightly lower MAE values compared to left-turns, suggesting greater model stability or predictability in through movements. These findings reinforce the advantage of GBBW model in achieving robust and accurate delay estimation across diverse traffic movement types and intersection configurations.

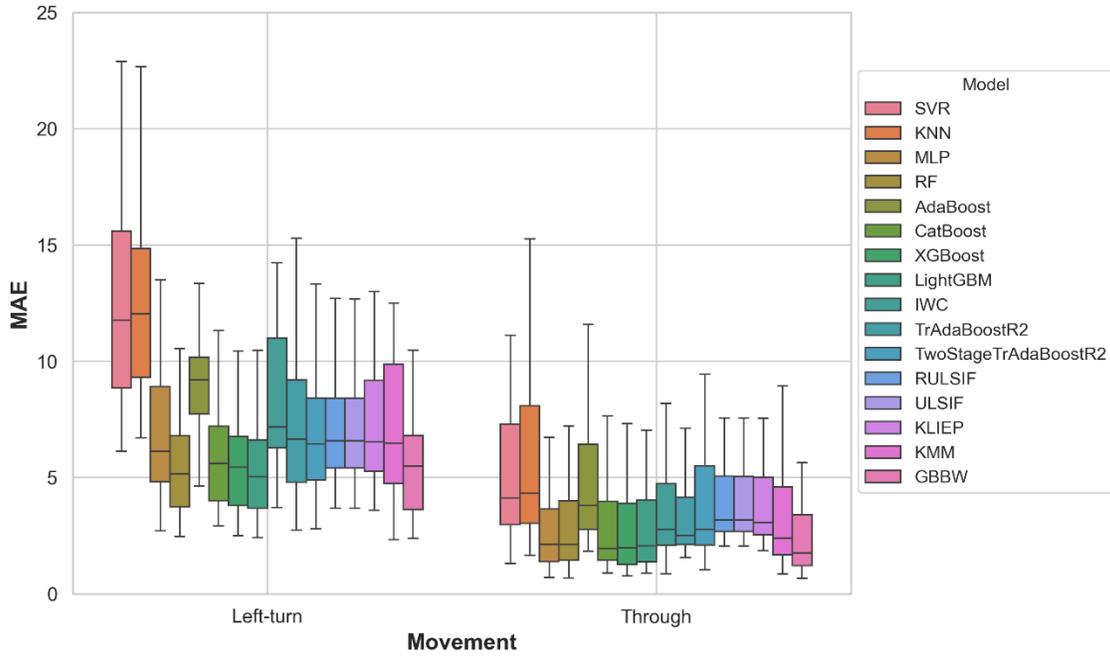

**Figure 4 MAE comparison across different models and movement types**

Figure 5 illustrates the RMSE distribution across multiple models for left-turn and through movements at signalized intersections. Traditional machine learning models such as SVR and KNN show the highest RMSE values, particularly for the left-turn movement, indicating lower accuracy and greater variability in estimating vehicle delay. In contrast, Boosting models like XGBoost, CatBoost, and LightGBM achieve noticeably lower RMSE values, especially for the through movement, suggesting improved performance. The proposed GBBW model consistently demonstrates competitive and superior performance, with lower median RMSE and narrower interquartile ranges in both movement types. The results reveal that the through movement tends to have lower RMSE across most models compared to the left-turn movement, suggesting that through movement delay is more predictable. Overall, the proposed GBBW model provides more robust and accurate estimations of vehicle delay, particularly in the presence of intersection heterogeneity.

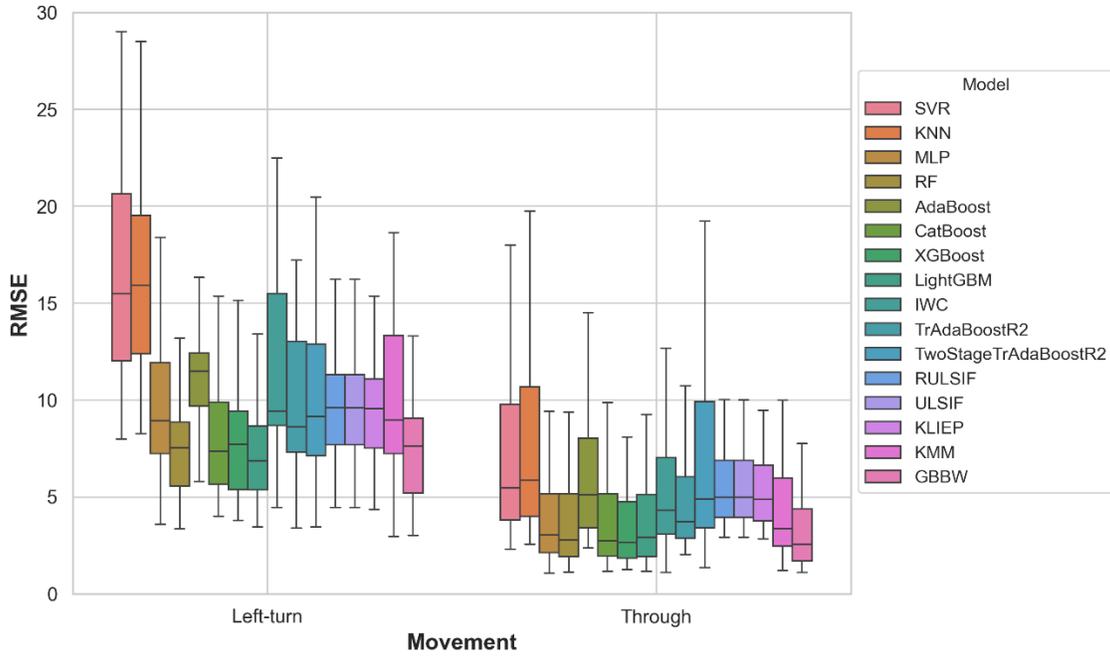

**Figure 5 RMSE comparison across different models and movement types**

### 3.6. Ablation Study on Data Samples

As shown in Figure 6, the ablation study reveals distinct trends in model performance for left-turn and through movements as the number of data samples increases. For the left-turn movement, MAPE shows a clear decline with increasing sample size, stabilizing at approximately 10.5% once the sample size reaches 72. This indicates that the model achieves consistent and reliable performance for the left turn movement with a sufficiently large dataset.

For the through movement, MAPE exhibits greater variability across sample sizes, but performance begins to stabilize at a sample size of 96, with fluctuations narrowing and error rates settling into a more predictable range. This suggests that while the through movement requires a larger dataset to achieve stability compared to the left turn movement, the model ultimately reaches a steady level of accuracy.

These findings highlight the importance of dataset size in traffic prediction models, with 72 samples representing a critical threshold for stability in left-turn movement predictions and 96 samples needed for through movement prediction. Beyond these sample sizes, additional data samples yield diminishing returns in error reduction, making them practical targets for efficient model training.

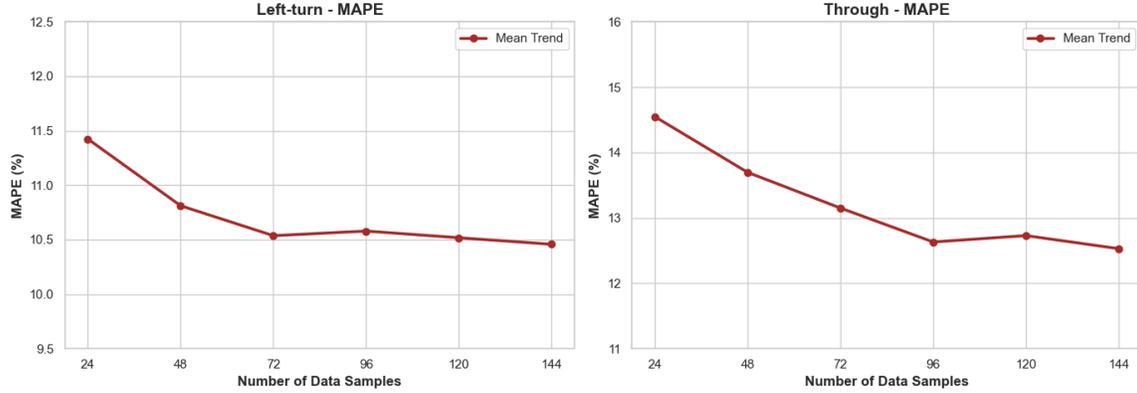

Figure 6 Average MAPEs across different sample sizes

## 4. CONCLUSION

Accurate vehicle delay estimation plays a vital role in evaluating and enhancing the operational efficiency of signalized intersections. As a direct indicator of traffic congestion and driver experience, delay impacts travel time reliability, fuel consumption, and emissions. In the context of modern and intelligent transportation systems, predictive or real-time delay estimation enables adaptive signal control, supports proactive congestion management, and informs performance-based infrastructure planning. Machine learning offers a scalable and cost-efficient alternative. Nonetheless, the effectiveness of conventional machine learning models is often limited by their reduced ability to generalize across intersections with diverse traffic patterns and operational characteristics.

To overcome this challenge, the proposed research presents a domain adaptation (DA) framework specifically tailored for vehicle delay estimation. By distinguishing between source and target domains and applying a fine-tuning step using limited labeled data from the target, the framework effectively adapts to diverse traffic patterns. The inclusion of the Gradient Boosting with Balanced Weighting (GBBW) model enhances adaptability by reweighting training instances based on similarity to the target domain, allowing the model to prioritize the most relevant data during training. This approach relaxes the assumption of identical data distributions and improves the model's applicability across heterogeneous intersections. Evaluation using data from 57 intersections in Pima County, Arizona, confirms the framework's robustness and flexibility.

Experimental results show that GBBW consistently outperforms both conventional machine learning and existing domain adaptation models in estimating vehicle delay. For both left-turn and through movements, GBBW achieved the lowest error metrics, highlighting its effectiveness in handling intersection-level heterogeneity. These findings demonstrate the proposed framework's potential to support data-driven traffic signal optimization and broader mobility planning, offering a scalable and reliable solution for transportation agencies.

Given the framework's high flexibility and capacity to incorporate multiple variables, future research could explore the inclusion of additional temporal and spatial factors to enhance model performance. Incorporating external influences such as weather conditions, special events, and traffic incidents may further strengthen the model's generalization capability across diverse traffic environments. As the first study to apply domain adaptation to vehicle delay estimation, this work lays the foundation for future advancements. Subsequent studies may consider integrating more advanced machine learning techniques and refined input features to further improve estimation accuracy and robustness.